\documentclass[conference]{IEEEtran}
\IEEEoverridecommandlockouts
\usepackage{cite}
\usepackage{amsmath,amssymb,amsfonts}
\usepackage{algorithmic}
\usepackage{graphicx}
\usepackage{textcomp}
\usepackage{xcolor}
\def\BibTeX{{\rm B\kern-.05em{\sc i\kern-.025em b}\kern-.08em
    T\kern-.1667em\lower.7ex\hbox{E}\kern-.125emX}}
\begin{document}

\title{Improving GNSS Positioning in Challenging Urban Areas by Digital Twin Database Correction\\

}

\author
{\IEEEauthorblockN{Jiarong Lian}
\IEEEauthorblockA{
\textit{Department of Aeronautical and }\\
\textit{Aviation Engineering}\\
\textit{The Hong Kong Polytechnic University}\\
Hong Kong\\
jiarong.lian@connect.polyu.hk}
\\
\IEEEauthorblockN{Guohao Zhang}
\IEEEauthorblockA{
\textit{Department of Aeronautical and }\\
\textit{Aviation Engineering}\\
\textit{The Hong Kong Polytechnic University}\\
Hong Kong\\
gh.zhang@polyu.edu.hk}
\and
\IEEEauthorblockN{Jiayi Zhou}
\IEEEauthorblockA{
\textit{Department of Aeronautical and }\\
\textit{Aviation Engineering}\\
\textit{The Hong Kong Polytechnic University}\\
Hong Kong\\
jiayi-louise.zhou@connect.polyu.hk}
\\
\IEEEauthorblockN{Li-Ta Hsu}
\IEEEauthorblockA{
\textit{Department of Aeronautical and }\\
\textit{Aviation Engineering}\\
\textit{The Hong Kong Polytechnic University}\\
Hong Kong\\
lt.hsu@polyu.edu.hk}
}

\maketitle

\begin{abstract}
Accurate positioning technology is the foundation for industry and business applications. Although indoor and outdoor positioning techniques have been well studied separately, positioning performance in the intermediate period of changing the positioning environment is still challenging. This paper proposed a digital twin-aided positioning correction method for seamless positioning focusing on improving the receiver's outdoor positioning performance in urban areas, where the change of the positioning environment usually happens. The proposed algorithm will simulate the positioning solution for virtual receivers in a grid-based digital twin. Based on the simulated positioning solutions, a statistical model will be used to study the positioning characteristics and generate a correction information database for real receivers to improve their positioning performance. This algorithm has a low computation load on the receiver side and does not require a specially designed antenna, making it implementable for small-sized devices.
\end{abstract}


\section{Introduction}
With the fast development of location-based services (LBS), accurate positioning and navigation become increasingly crucial. In civil applications, positioning technologies can be classified into two groups, outdoor positioning technologies and indoor positioning technologies. For indoor positioning technologies, positioning can be achieved by ultra-wideband (UWB), Wireless Fidelity (Wi-Fi) round-trip-time (RTT), cellular, etc. Among these technologies, UWB and Wi-Fi RTT are the most popular. UWB can achieve 10-centimeter level accuracy \cite{10.1145/3448303} by lateration techniques (time of arrival (TOA), time difference of arrival (TDOA)), Wi-Fi RTT can achieve meter-level accuracy \cite{9110232,105208911111111}. For outdoor positioning technologies, positioning can be achieved by cellular, Wi-Fi, inertial navigation system (INS), global navigation satellite system (GNSS), etc. GNSS is the most widely used outdoor positioning technology because it is widely covered and capable of providing an absolute positioning solution. In the open-sky environment, GNSS can achieve meter-level accuracy \cite{10.1007/s10291-018-0736-8}.

Although the above-mentioned technologies can achieve satisfying positioning accuracy in indoor and outdoor environments accordingly, the positioning performance of the intermediate period for switching the environment (from indoor environment to outdoor environment or from outdoor environment to indoor environment) in urban areas is not satisfied. In this scenario, signals of the indoor technologies will be significantly affected by the building walls. The building surfaces may also block or reflect the GNSS signal, causing more than 50 meters of positioning error \cite{RN2}. There are two types of undesired GNSS signal receptions in urban areas: multipath effects and non-line-of-sight (NLOS) receptions. The former refers to the scenario in which the receiver receives both direct and reflected GNSS signals and the latter refers to the scenario in which the receiver receives solely the reflected GNSS signal. These two types of GNSS signal reception will result in inaccurate pseudorange measurements, leading to poor GNSS positioning performance in urban areas. NLOS reception and multipath effect are illustrated in Fig. 1.

\begin{figure}[tb]
\centerline{\includegraphics[scale = 0.8]{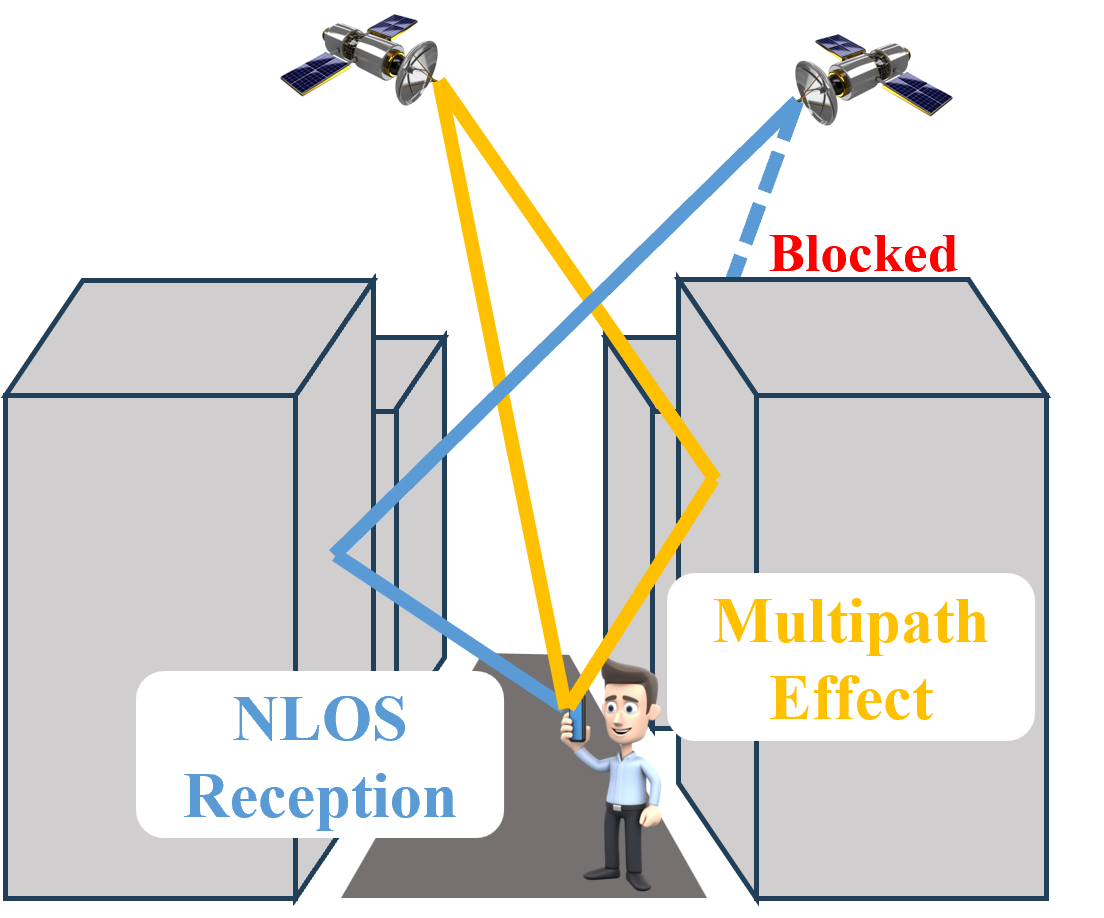}}
\caption{NLOS reception (in blue) and multipath effect (in yellow).}
\end{figure}
\begin{figure*}[t]
\centerline{\includegraphics[scale = 0.8]{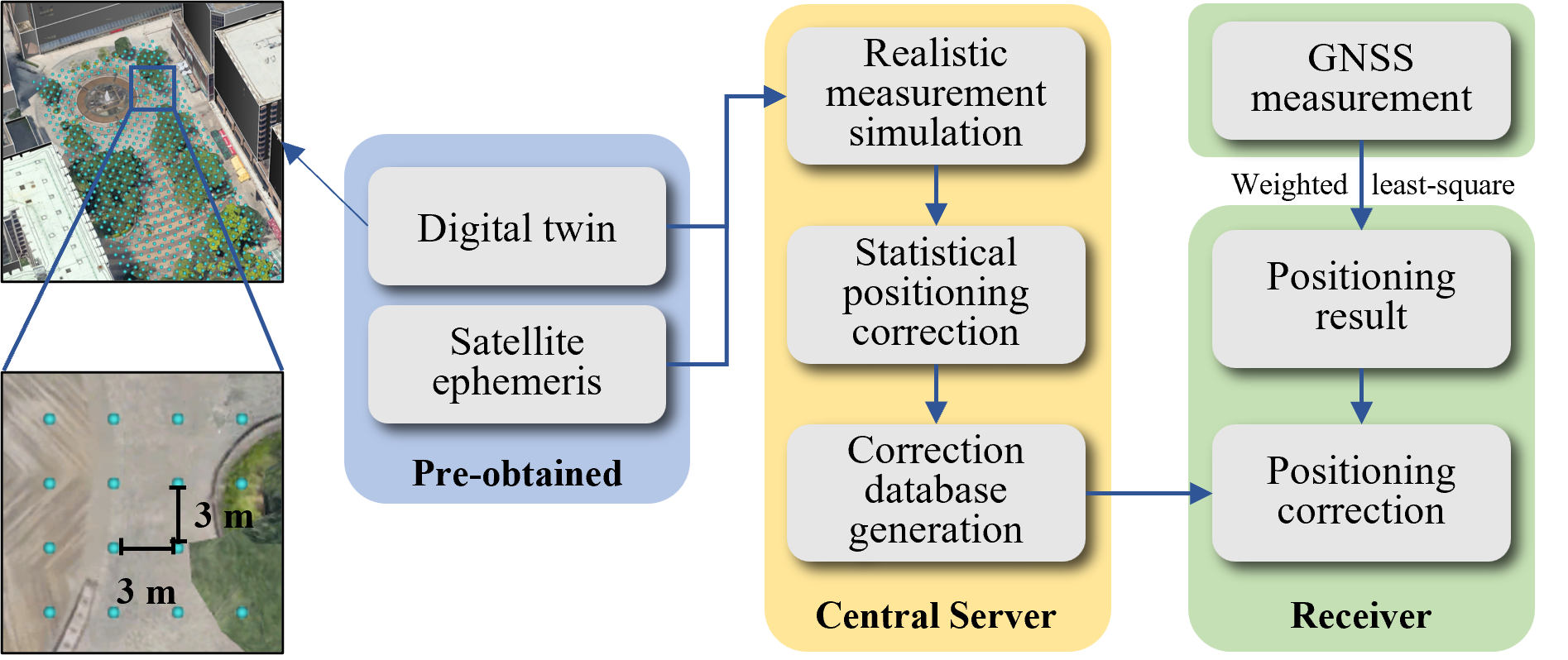}}
\caption{System structure of the proposed DT-aided GNSS positioning algorithm.}
\end{figure*}
Numerous approaches have been proposed to mitigate the multipath effect and NLOS receptions for better GNSS urban positioning performance. These approaches can be divided into three categories: 1) Antenna-level improvement. 2) Signal-level improvement. 3) Positioning-level improvement. Antenna-level improvement focuses on physically mitigating GNSS signals with multipath and NLOS effects. For example, a chock-ring designed GNSS antenna can mitigate low elevation GNSS signals, which are more likely to be reflected GNSS signals \cite{299591}. Signal-level improvement focuses on mitigating reflected GNSS signals by detecting their special features. For example, de-weighting GNSS measurements with a lower carrier-to-noise ratio (\(C/N_{0}\)) can mitigate multipath and NLOS effect in positioning because reflected GNSS signals will be attenuated by the reflection surfaces \cite{RN2}. However, these two levels of improvement are hard to implement in some civil applications as the former needs a large antenna and the latter needs raw GNSS measurement. Considering these factors, positioning-level improvement is a good way to improve the positioning accuracy for seamless positioning.

Previous studies have shown a lot of potential of digital twin (DT) to solve positioning and wireless communication problems \cite{s22165936,Addonavi.644,9429703}. As a digital version of the real world, DT enables computers to compute and simulate the GNSS signal propagation in the real world. Existing works on 3D mapping-aided (3DMA) GNSS can be regarded as one of the approaches in DT-aided GNSS \cite{10.1007/s10291-015-0451-7,9082815,Wang_Groves_Ziebart_2015}. These algorithms improve urban positioning accuracy by matching the receiver's satellite visibility or pseudorange measurement with the computer-simulated result generated from the city 3D map and ray-tracing \cite{Zhongnavi.515,https://doi.org/10.1002/navi.85,10491363}. However, these algorithms require a huge computational load on the receiver side, making them hard to implement. 

To solve this problem, a DT-aided GNSS positioning algorithm is proposed in this study. This positioning-level improvement aims to simulate the receivers' positioning performance on the DT and generate correction information for receivers in the real world. DT-aided GNSS will realistically simulate the receiver's positioning errors that may happen in the real world due to the NLOS effect and statistically generate a positioning correction database. All the calculations can be implemented on a central server and sent to the receivers in the real world. They can use the generated correction information to improve their positioning accuracy. The contributions of this study are:

\begin{enumerate}
\item A novel DT simulator is developed to simulate the GNSS pseudorange measurement and positioning error in urban areas.
\item A statistical model is proposed to correct the real-world positioning solution based on the DT simulation.
\item An experiment is carried out to assess the performance of the proposed algorithm.
\end{enumerate}

\section{Methodology}
\subsection{System Structure}
The overall system structure is illustrated in Fig. 2. There are two kinds of pre-obtained data for the simulation, one is the DT of the simulated area and another is the satellite ephemeris. DT is a digital version of the real world that enables the computer to simulate the GNSS signal propagation based on known physics knowledge. The DT used in this paper is a grid-based DT, which separates the DT into several 3-meter by 3-meter grids. The two figures on the left side of Fig. 2 illustrate this gridding idea. The satellite ephemeris is obtained from the Lands Department of the Hong Kong Government \cite{RN36}. The central server will utilize the pre-obtained data to simulate the receiver's pseudorange measurement and expected positioning solution at each grid inside a grid-based DT using the ray-tracing (RT) technique. After simulating the receiver's positioning solution at all grids for a certain period of time, a statistical analysis will be carried out to model the positioning performance in this area. A correction information database can be generated based on the statistical model of the simulated receiver's positioning solution in each grid. This correction information database will be transferred to the receiver via the Internet. After the receiver receives the GNSS signals and calculates its positioning solution in the real world, the correction information database will help the receiver correct its positioning solution for better positioning accuracy.

\subsection{Ray-tracing Simulation}
Based on the building surfaces in DT and the satellite ephemeris, the GNSS signal propagation path can be simulated \cite{rs13040544}. In the realistic measurement simulation, RT will be used to determine the GNSS signal propagation path. This simulator assumes the signal will have a standard spectrum reflection on the building surface and can only reach the receiver without reflection or with one reflection. Based on this assumption, the total travel distance of the reflected GNSS signal can be obtained by calculating the range from the satellite to the point that mirrors the receiver to the building surface. There are two kinds of valid GNSS signal reception. If the signal can propagate to the receiver directly without any reflection, it belongs to the line-of-sight (LOS) reception. If the signal is reflected to the receiver by the building surfaces, it belongs to the NLOS reception. These two kinds of reception are depicted in Fig. 3. 

\begin{figure}[ht]
\centerline{\includegraphics[scale=0.72]{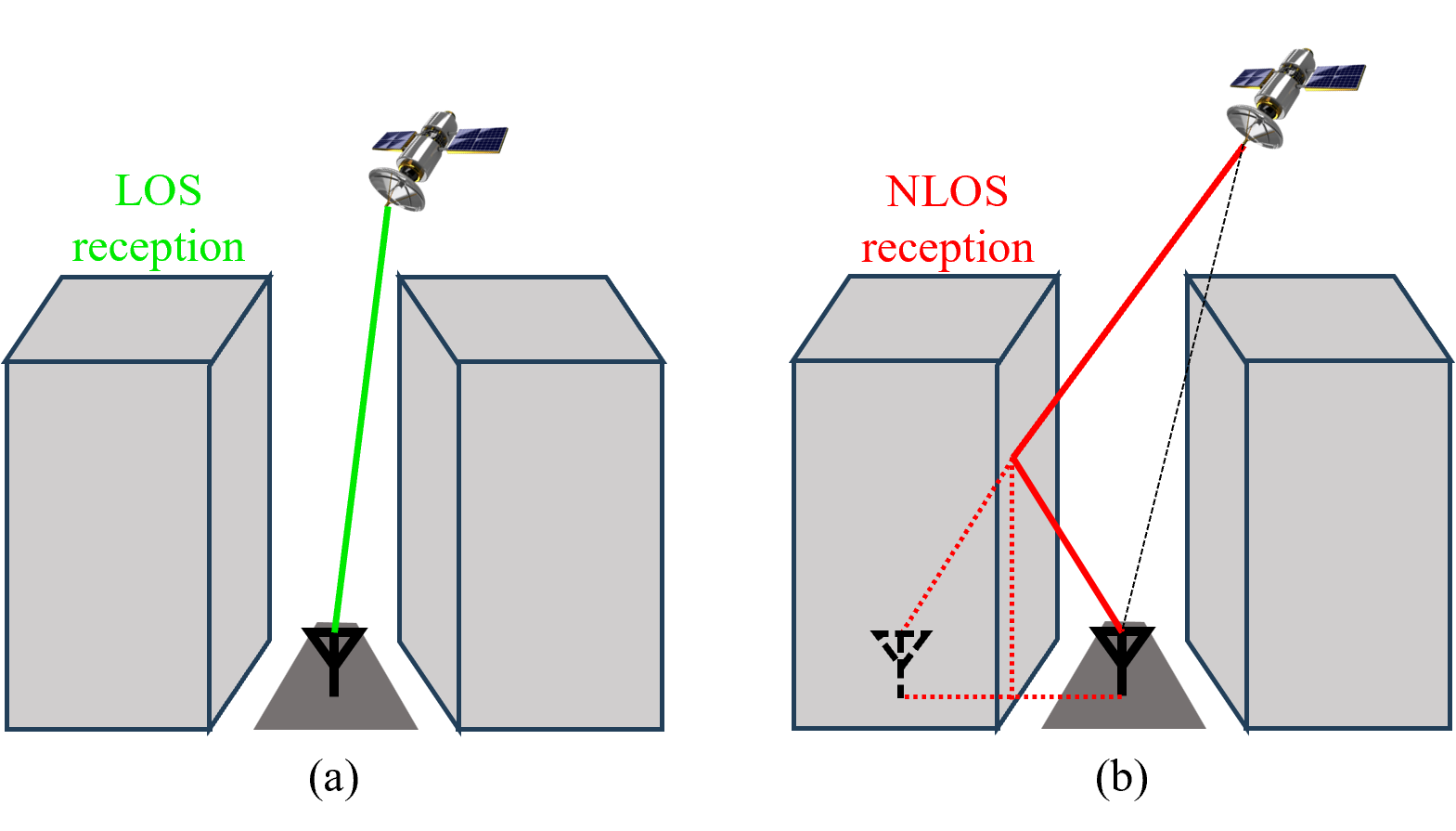}}
\caption{Valid GNSS signal reception in the realistic measurement simulation. (a) LOS reception. (b) NLOS reception.}
\end{figure}

Virtual receivers can be generated for every grid inside the grid-based DT. For grid indexed \(k\), the virtual receiver \(\textbf{x}_{k,DT}\) is located at the center of each grid with a height of one meter above the ground assuming the simulation result caused by a small difference on height is negligible. The ground altitude is obtained from the Hong Kong digital terrain model provided by the Land Department of the Hong Kong Government \cite{RN25}. A realistic measurement simulator will simulate the pseudorange measurement of all virtual receivers in DT. The simulated pseudorange measurement can be expressed as:

\begin{equation}
\rho^{sv}_k =
\begin{cases}
r^{sv}_k+\varepsilon & \text{LOS reception} \\
r^{sv}_k+d^{sv}_k+\varepsilon & \text{NLOS reception}
\end{cases}
\end{equation}
where \(\rho^{sv}_k\) denotes the simulated pseudorange measurement of the satellite \(sv\) by the virtual receiver \(\textbf{x}_{k,DT}\), \(r^{sv}_k\) denotes the range from the satellite \(sv\) to the virtual receiver \(\textbf{x}_{k,DT}\), \(d^{sv}_k\) denotes the extra travel distance of GNSS signal due to the reflection, and \(\varepsilon\) denotes the pseudorange measurement bias modeled by the atmospheric delay, clock offset, etc.

Based on the simulated pseudorange measurement, the simulated positioning solution for the virtual receiver \(\textbf{x}_{k,DT}\) can be calculated by the ordinary least-squares (OLS) estimation in a matrix form:

\begin{equation}
\mathbf{\Delta \mathbf{x}} = (\textbf{G}^\intercal \textbf{G})^{-1}\textbf{G}^\intercal\textbf{b}
\end{equation}

\begin{equation}
\mathbf{\Delta \mathbf{x}} = 
{\left[ \begin{array}{c}
\Delta x\\
\Delta y\\
\Delta z\\
c\Delta t
\end{array}
\right ]}
\end{equation}
\begin{equation}
\textbf{G} = 
{\left[ \begin{array}{cccc}
\frac{(x^1-x_i)}{\rho^1_0} & \frac{(y^1-x_i)}{\rho^1_0} & \frac{(z^1-x_i)}{\rho^1_0} & 1\\
\frac{(x^2-x_i)}{\rho^2_0} & \frac{(y^2-x_i)}{\rho^2_0} & \frac{(z^2-x_i)}{\rho^2_0} & 1\\
... & ... & ... & ...\\
\frac{(x^{sv}-x_i)}{\rho^{sv}_0} & \frac{(y^{sv}-x_i)}{\rho^{sv}_0} & \frac{(z^{sv}-x_i)}{\rho^{sv}_0} & 1
\end{array}
\right ]}
\end{equation}
\begin{equation}
\textbf{b} =
{\left[ \begin{array}{cccc}
\rho^1_i-\rho^1_{i-1}\\
\rho^2_i-\rho^2_{i-1}\\
... \\
\rho^{sv}_i-\rho^{sv}_{i-1}
\end{array}
\right ]}
\end{equation}
where \(\mathbf{\Delta \mathbf{x}}\) denotes the estimated state vector and time bias that consist of the initial prediction and the increment from the estimation. \textbf{G} denotes the geometrical matrix incorporating unit LOS vectors. \textbf{b} denotes the difference between the estimation predicted pseudorange measurement and the simulator-estimated pseudorange measurement.

\subsection{Positioning Correction from Digital Twin}
After applying the ray-tracing simulation, all potential GNSS positioning outcomes at different virtual receiver locations within a certain area have been predicted in the DT. Each candidate receiver location can be associated with a wrongly estimated location due to NLOS errors. Thus, the original receiver location associated with the estimated solution in a real application can be retrieved from the DT, potentially indicating the receiver’s true location. However, such association is not distinct, i.e., multiple candidate receiver locations can have the biased measurement lead to the same incorrect positioning result. In this study, the positioning correction is estimated statistically, as demonstrated in Fig. 4.

A certain area in the digital twin will be grid into candidate receiver positions with a resolution of three meters. For the \(k^{th}\) original candidate position \(\textbf{x}_k\)  at epoch \(t\), its associated incorrect positioning solution can be obtained from the digital twin, by applying the OLS estimation with the ray-tracing simulated measurements, as follows:

\begin{equation}
\widehat{\textbf{x}}^t_{k,DT} = \textbf{x}_k + \boldsymbol\varepsilon^t_{k,DT}
\end{equation}
where \(\boldsymbol\varepsilon^t_{k,DT}\) denotes the vector of positioning error introduced by the NLOS errors (related to the satellite positions at \(t\)) predicted at the candidate position \(\textbf{x}_k\). 

On the other hand, the same set of grid points can be used to represent candidate resulting positions, denoted as \(\textbf{x}_n\)  with the index \(n\). Then, different \(\textbf{x}_k\)  may have the same biased solution resulting in \(\textbf{x}_n\), which can be collected as a set, using:

\begin{equation}
\{\textbf{x}_{k'}|\text{ }\widehat{\textbf{x}}^t_{k',DT} = \textbf{x}_n,\text{ } k' \epsilon k\}
\end{equation}
where \(k'\) is the index of original candidates with a biased solution to \(\textbf{x}_n\) due to the error \(\boldsymbol\varepsilon^t_{k',DT}\). In this study, to consider a certain level of noise in the time domain, the 24 hours of ephemeris data will be divided into different slots with 300 seconds, to apply digital twin simulation representing the phenomenon within each time slot. Thus, the resulting candidates within 300 seconds of the time slot corresponding to epoch t will be considered in equation (7). In practice, the digital twin simulation can be extended to different days according to the periodicity of satellite positions \cite{10012445}.

Each \(\widehat{\textbf{x}}^t_{k',DT}\) can be corrected by applying a compensation of \(\boldsymbol\varepsilon^t_{k',DT}\). For a specific \(\textbf{x}_n\)  with multiple \(\text{ }\widehat{\textbf{x}}^t_{k',DT}\) from different original position \(\textbf{x}_{k'}\) , the overall correction can be estimated by:

\begin{equation}
\Delta \textbf{x}_n = - \sum_{k'=1}^{K'} \frac{\boldsymbol\varepsilon^t_{k',DT}}{K'}
\end{equation}
where \(N'\) is the total number of candidates indexed by \(N'\). Thus, in real applications, for a receiver with the initial position measured as \(\widetilde{\textbf{x}}_k\), the receiver position can be rectified by:

\begin{equation}
{\widehat{\textbf{x}}_{DT}={\widetilde{\textbf{x}}}_n + \Delta \textbf{x}_n}
\end{equation}

\begin{figure}
    \centering
    \includegraphics[scale=0.58]{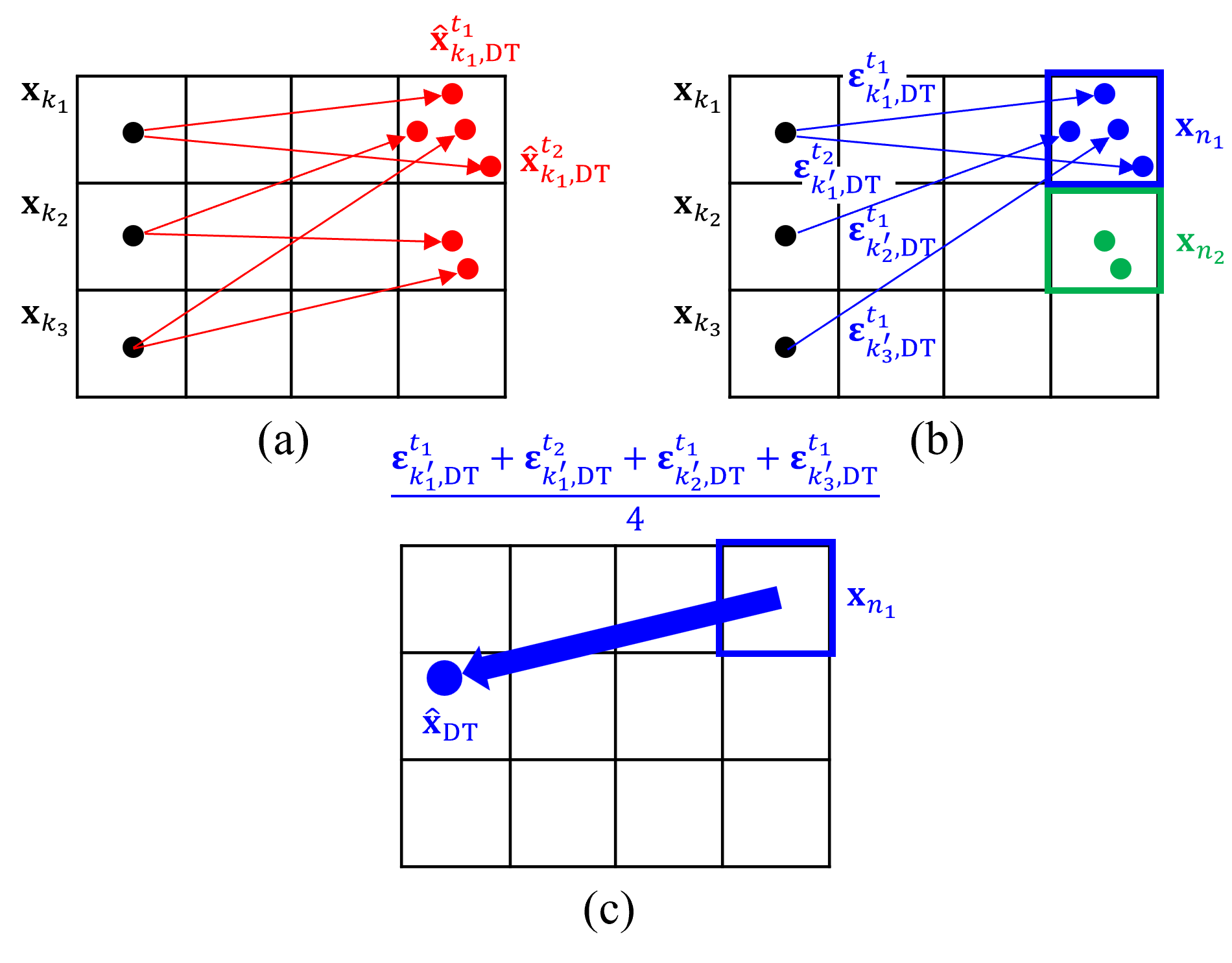}
    \caption{Demonstration of position correction estimated from digital twin database. (a) Positioning error simulation of original candidates; (b) Collecting solutions biased to the same candidate according to digital twin; (c) Estimation of the overall correction vector corresponding to a particular resulting candidate position. }
    \label{fig:enter-label}
\end{figure}





\section{Experiment Evaluation}

\subsection{Experiment Setup}

The configuration of the experiment is shown in Fig. 5. A commercial GNSS receiver (Ublox F9P) is attached to the researcher's head and connected to the computer by a wire. This configuration avoids unwanted GNSS signal interference from moving vehicles and pedestrians. There is also a camera attached to the researcher's head. The recorded video will be used to determine the ground truth by matching the notable buildings with Google Earth Pro.

\begin{figure}[ht]
\centerline{\includegraphics[scale=1.2]{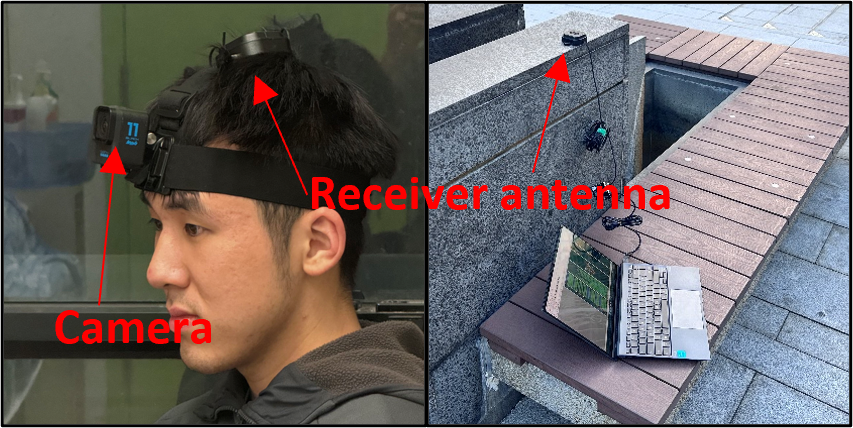}}
\caption{Experiment setup. GNSS antenna is attached to the researcher's head. A camera is also attached to record ground truth information.}
\end{figure}

The experiment is carried out on a narrow street in Tsim Sha Tsui, Hong Kong, which is shown in Fig. 6. The researcher is walking on the sidewalk from the right side to the left side, assessing the positioning performance of the proposed algorithm for the intermediate period that the receiver is changing from outdoor environment to indoor environment. The update rate of the GNSS receiver is 1 Hz and the duration of this experiment is 63 seconds, so there are 63 positioning solutions available in this experiment to evaluate the performance of the proposed algorithm. 

There are two kinds of experimental results being evaluated to assess the positioning performance of the proposed algorithm, which are:
\begin{itemize}
\item \textbf{Weighted Least-squares (WLS)}: This positioning result is calculated by applying WLS to the receiver's pseudorange measurement. This algorithm is widely used and usually integrated into low-cost GNSS chips \cite{Realini_2013}, so it is selected to represent the receiver's normal performance (without correction).
\item \textbf{DT-aided GNSS}: This positioning result is calculated by applying the proposed algorithm to the WLS positioning solution (with correction).
\end{itemize}

\begin{figure}[ht]
\centerline{\includegraphics[scale=0.55]{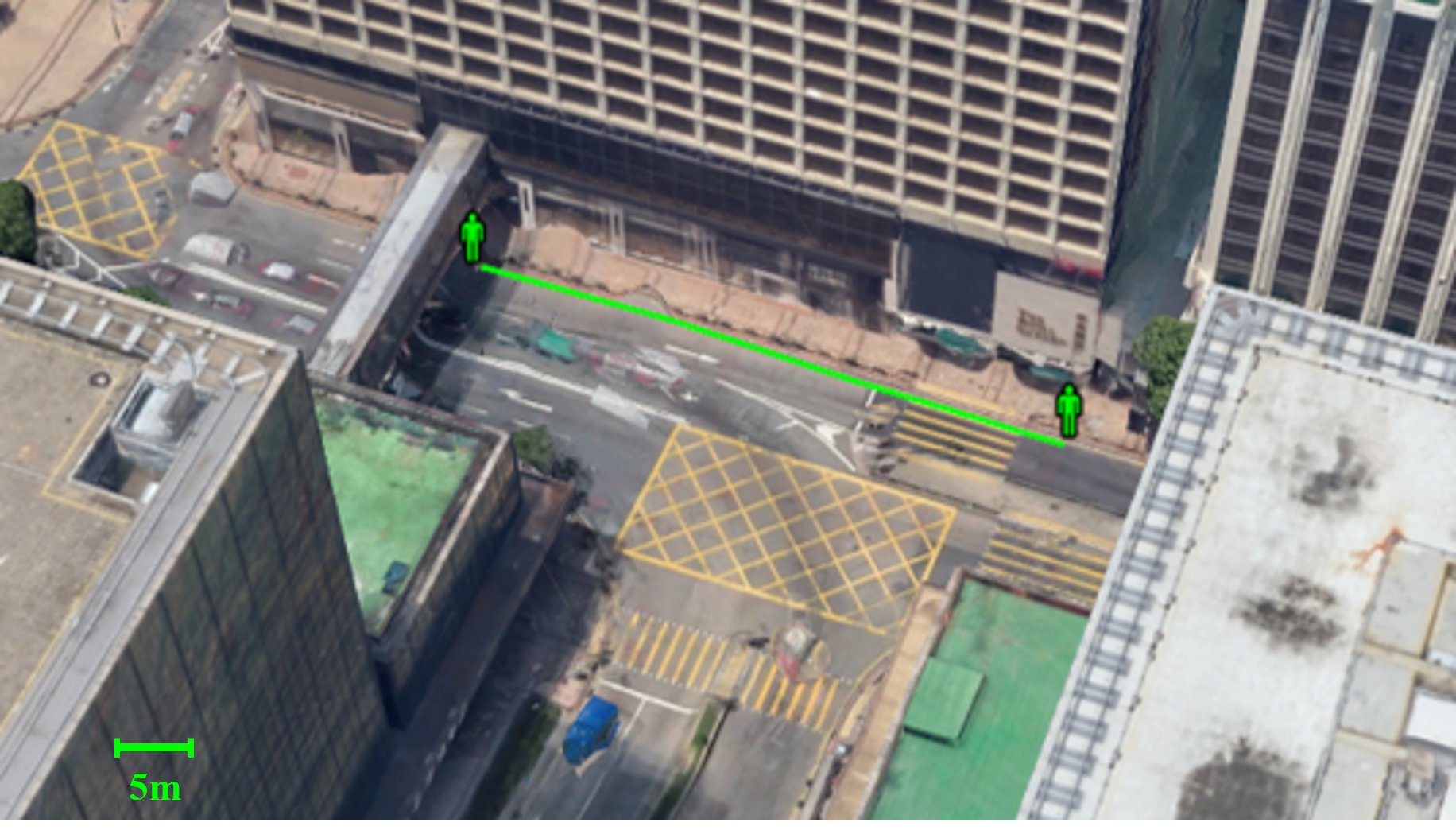}}
\caption{Experiment environment. The researcher is walking from the right side to the left side.}
\end{figure}

\subsection{Experiment Result}

A comparison of the results of the two algorithms in a 2D map and the line graph of the 2D positioning error is illustrated in Fig. 7 and Fig. 8. A detailed 2D positioning error statistics of the WLS positioning algorithm and DT-aided GNSS positioning algorithm are shown in Table I. The mean of the 2D positioning error of the proposed DT-aided GNSS is 19.8 meters, which reduces more than 50\% of the 2D positioning error compared to 41.2 meters of the WLS positioning algorithm. At the same time, DT-aided GNSS reduces the root mean square of the 2D positioning error from 42.1 meters to 23.4 meters compared to WLS, proving the effectiveness of DT-aided GNSS in improving positioning performance in urban areas. This positioning performance improvement can also be identified from the map that the positioning result of DT-aided GNSS lies on the correct side of the street for most of the experiment results. 

Fig. 9 illustrates how the proposed DT-aided GNSS positioning algorithm works in a real application. At this moment (the \(22^{nd}\) second), the estimated positioning solution from the WLS positioning algorithm based on real pseudorange measurement is located at the lower part of the map. In the DT simulation, there are four virtual receivers that have the simulated positioning bias pointing to this position. The relative number of the estimated positioning bias pointing to this positioning for each virtual receiver is reflected in the color of each virtual receiver in the figure. For these virtual receivers, three out of four are close to the ground truth while the only outlier has a lower number of estimated positioning biases pointing to the WLS positioning solution. Thus, the estimated final state of DT-aided GNSS is close to the ground truth. These results reflect that 1) the realistic measurement simulation has successfully simulated the receiver's positioning solution and 2) the statistical model has successfully corrected the receiver's positioning solution. 

However, the proposed algorithm is not stable at a certain time, for example, time stamps of the \(34^{th}\) second, the \(36^{th}\) second, and the \(55^{th}\) second in this experiment. At these time epochs, the 2D positioning error goes up dramatically and is the same with the WLS positioning solution. The reason behind this is that at these time epochs, the positioning solution of real-world GNSS pseudorange measurement is located at the area where DT-aided GNSS does not have any correction information due to some unmodelled bias. This is the limitation of the proposed algorithm because the correction information is generated based on the simulated positioning solution and the simulated positioning solution cannot reach every location in the area. At the same time, the standard deviation of the 2D positioning error of the proposed algorithm is 12.7 meters, which is higher than the 9.1 meters of the WLS positioning algorithm, indicating that this algorithm is not smooth and robust. A possible way to improve this is to add additional information to the algorithm using a filter. These issues can be addressed in a future study.

\begin{table}[htbp]
\caption{Positioning Error (2D) Statistics (meters)}
\begin{center}
\begin{tabular}{c c c c c c}
\hline
\textbf{Algorithm}&\textbf{Mean}&\textbf{STD}&\textbf{RMS}&\textbf{Max}&\textbf{Min}\\
\hline
WLS&41.2&9.1&42.1&63.1&17.3\\
\hline
DT-aided GNSS&19.8&12.7&23.4&62.6&2.3\\
\hline
\end{tabular}
\end{center}
\end{table}

\begin{figure}[htbp]
\centerline{\includegraphics[scale=0.72]{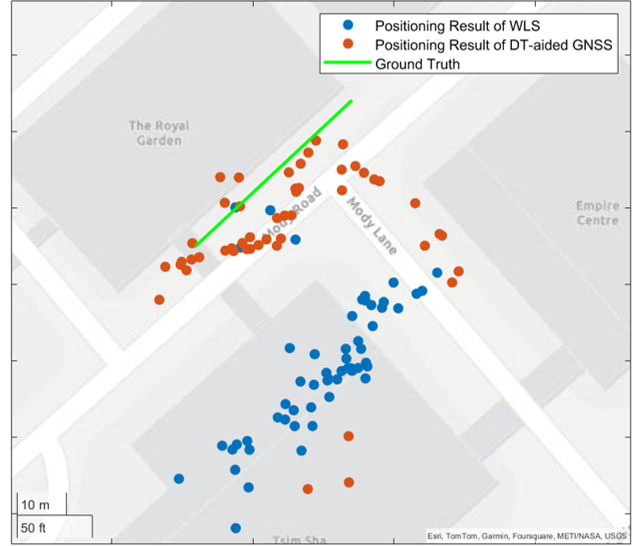}}\caption{Graphical experiment result in a 2D map.}
\end{figure}

\begin{figure}[htbp]
\centerline{\includegraphics[scale=1.57]{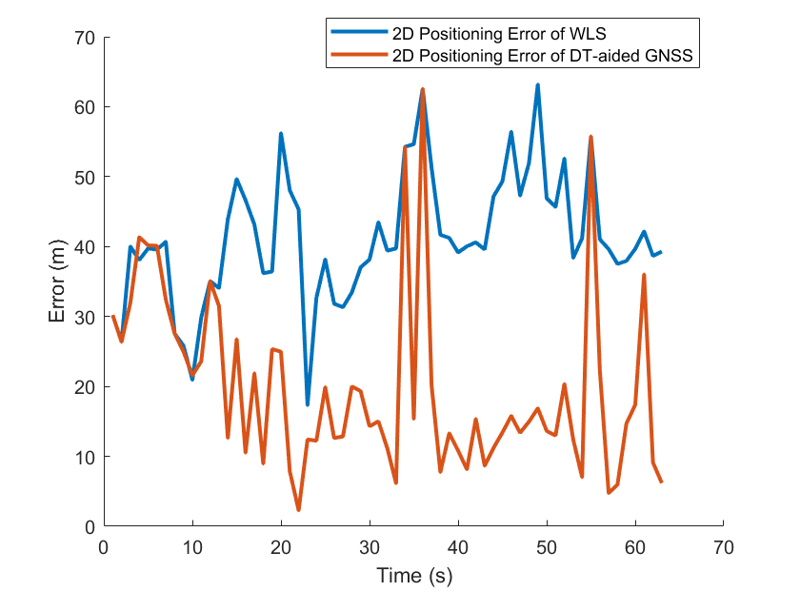}}\caption{Line graph of the 2D positioning error of WLS positioning algorithm (in blue) and DT-aided GNSS positioning algorithm (in orange).}
\end{figure}

\begin{figure}[htbp]
\centerline{\includegraphics[scale=0.85]{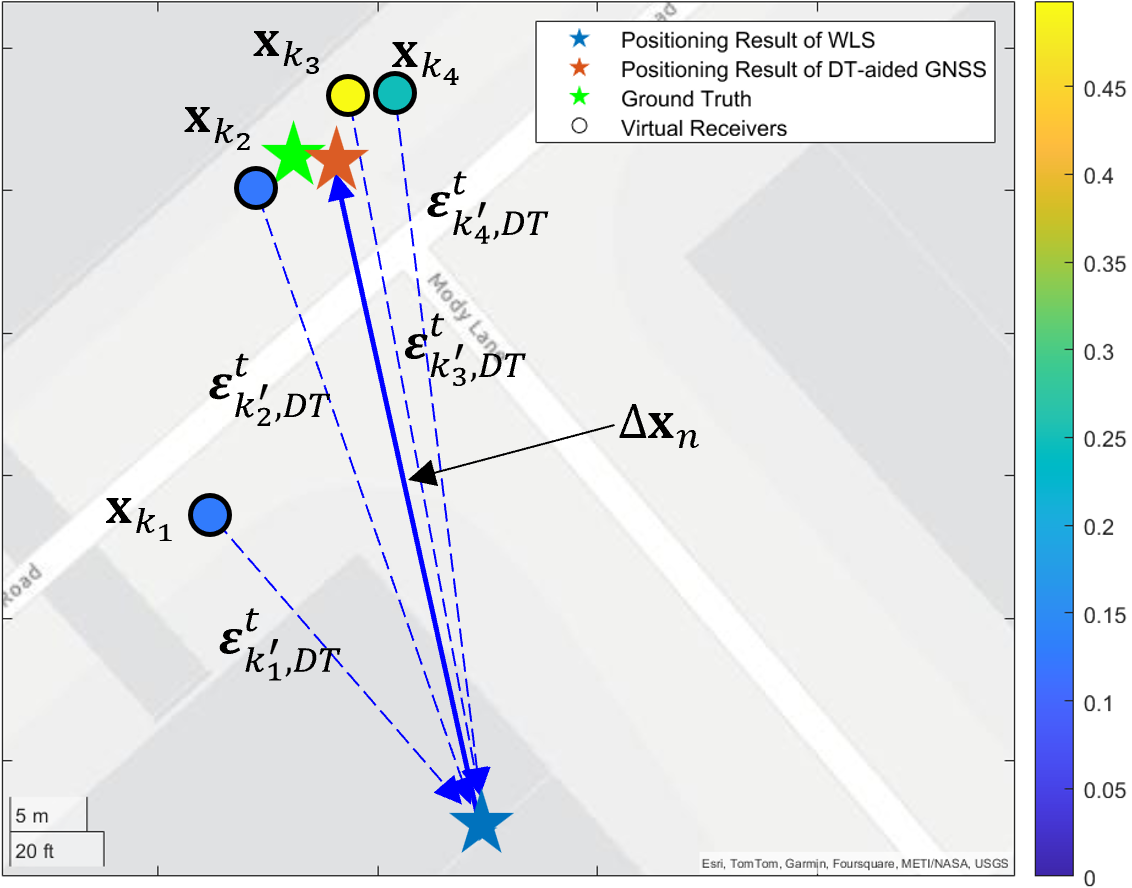}}\caption{Detailed DT-aided GNSS positioning algorithm verification at the \(22^{nd}\) second. The color-bar represents the weighted number of simulated biases pointing to the WLS positioning solution.}
\end{figure}

\section{Conclusion and Future Work}

This paper proposed a novel DT-aided GNSS positioning method for seamless positioning, which especially focuses on improving the intermediate period when the receiver changes the positioning environment. To reduce the computation load on the receiver's side, this algorithm simulates the positioning solution of different locations in DT by introducing virtual receivers. A statistical model is used to process the simulation result and generate correction information for receivers to improve its positioning solution in the real world. We evaluate the positioning performance of the proposed algorithm in an urban area, it reduces more than 50\% of the 2D positioning error compared to WLS and identifies the correct side of the street the researcher is walking on. However, there are still some limitations that need to be solved. This algorithm cannot provide correction information to all the locations due to some unmodelled bias, and its positioning solution is not smooth and robust. In future work, we will consider using the Gaussian progress regression to estimate the correction information for the locations that do not have correction information in this algorithm based on the known correction information. Bayesian filters will also be considered to improve the system's smoothness. At the same time, improving the accuracy of RT simulation can enhance the performance of DT-aided GNSS. For example, positioning error caused by the receiver's surrounding dynamical objects (vehicles, pedestrians, etc.) can be considered in the RT simulation. An efficient RT simulation strategy will be explored to lower the computation load in the server, which can make DT-aided GNSS suitable for large scale applicaitons.

\section*{Acknowledgment}
This work was substantially supported by the National Natural Science Foundation of China (Grant No. 62303391) and partly by PolyU Departmental Start-up Fund entitled “Outdoor/Indoor Seamless Navigation by a 3D Mapping Aided GNSS/Wi-Fi Integrated Positioning System” under P0043807.
\bibliographystyle{IEEEtran}
\bibliography{bib.bib}

\end{document}